\documentclass[twocolumn]{article}
\usepackage{spconf,amsmath,graphicx}
\usepackage[noend]{algpseudocode}
\usepackage{algorithm,algorithmicx}
\usepackage{makecell}
\usepackage{float}
\usepackage{stfloats}
\usepackage{booktabs}

\usepackage[colorlinks=true,citecolor=green]{hyperref}
\usepackage{cleveref}
\usepackage{tabularx}
\usepackage{caption}
\usepackage{subcaption}
\usepackage{adjustbox}
\usepackage{subfiles}
\usepackage{amssymb}
\usepackage{pifont}
\usepackage{multirow}
\usepackage{xcolor}
\usepackage{ieeetrantools}
\usepackage{cite}
\usepackage{tikz}

\newcommand*\Let[2]{\State #1 $\gets$ #2}

\makeatletter
\renewcommand\section{\@startsection {section}{1}{\z@}%
                                  {-2.5ex \@plus -1ex \@minus -.2ex}%
                                  {1.75ex \@plus.2ex}%
                      {\normalfont\Large\bfseries}}
\renewcommand\subsection{\@startsection{subsection}{2}{\z@}%
                                     {-1.5ex\@plus -1ex \@minus -.2ex}%
                                     {1.ex \@plus .2ex}%
                              {\normalfont\large\bfseries}}
\renewcommand\subsubsection{\@startsection{subsubsection}{3}{\z@}%
                                     {-3.25ex\@plus -1ex \@minus -.2ex}%
                                     {1.5ex \@plus .2ex}%
                          {\normalfont\normalsize\bfseries}}
\renewcommand\paragraph{\@startsection{paragraph}{4}{\z@}%
                                    {3.25ex \@plus1ex \@minus.2ex}%
                                    {-1em}%
                        {\normalfont\normalsize\bfseries}}
\renewcommand\subparagraph{\@startsection{subparagraph}{5}{\parindent}%
                                    {3.25ex \@plus1ex \@minus .2ex}%
                                    {-1em}%
                          {\normalfont\normalsize\bfseries}}
\makeatother

\title{One-Shot Sensitivity-Aware Mixed Sparsity Pruning\\ for Large Language Models}
\name{Hang Shao$^1$, Bei Liu$^1$, Bo Xiao$^2$, Ke Zeng$^2$, Guanglu Wan$^2$, Yanmin Qian$^1$\textsuperscript{\dag} \thanks{\textsuperscript{\dag}  
corresponding author
}}
\address{$^1$Auditory Cognition and Computational Acoustics Lab\\
MoE Key Lab of Artificial Intelligence, AI Institute\\
Department of Computer Science and Engineering, Shanghai Jiao Tong University, Shanghai, China\\$^2$Meituan, Beijing, China\\}

\begin{document}
\bstctlcite{IEEEexample:BSTcontrol} 
\ninept

\maketitle
\begin{abstract}
Various Large Language Models~(LLMs) from the Generative Pretrained Transformer~(GPT) family have achieved outstanding performances in a wide range of text generation tasks. However, the enormous model sizes have hindered their practical use in real-world applications due to high inference latency. Therefore, improving the efficiencies of LLMs through quantization, pruning, and other means has been a key issue in LLM studies. In this work, we propose a method based on Hessian sensitivity-aware mixed sparsity pruning to prune LLMs to at least 50\% sparsity without the need of any retraining. It allocates sparsity adaptively based on sensitivity, allowing us to reduce pruning-induced error while maintaining the overall sparsity level. The advantages of the proposed method exhibit even more when the sparsity is extremely high. Furthermore, our method is compatible with quantization, enabling further compression of LLMs. We have released the available code\footnote{https://github.com/talkking/MixGPT}.

\end{abstract}


\section{Introduction}
\label{sec:introduction}


Large language models (LLMs) from the GPT family have demonstrated exceptional performances across a wide range of tasks. However, due to their large sizes and high computational costs, model deployment face new challenges upon GPU memory consumption as well as inference latency.
Consequently, there's a prominent need to compress LLM models to a feasible range in order to make use of them in real applications.
In the literature, there have been various mainstream model compression techniques available, for pre-trained models in particular, including knowledge distillation\cite{Sun_Gan_Fang_Cheng_Wang_Liu_2020,Pan_Wang_Qiu_Zhang_Li_Huang_2021,Sun_Cheng_Gan_Liu_2019,sanh2019distilbert}, model quantization\cite{Bai_Zhang_Hou_Shang_Jin_Jiang_Liu_Lyu_King_2021,Frantar_Ashkboos_Hoefler_Alistarh_2022,Yao_Aminabadi_Zhang_Wu_Li_He_2022,Xiao_Lin_Seznec_Demouth_Han_2022,Dettmers_Svirschevski_Egiazarian_Kuznedelev_Frantar_Ashkboos_Borzunov_Hoefler_Alistarh_2023}, model sparsity pruning\cite{Wang_Wohlwend_Lei_2020,Kurtic_Campos_Nguyen_Frantar_Kurtz_Fineran_Goin_Alistarh,Zafrir_Larey_Boudoukh_Shen_Wasserblat,Ma_Fang_Wang,Frantar_Alistarh_2023}, and \textit{etc}.

Knowledge distillation focuses on transferring knowledge from a large teacher model to a smaller student model by predicting pseudo-labels or leveraging the knowledge from intermediate hidden layers. This process enables the compact student model to achieve approximate accuracy comparable to that of the teacher model. Model quantization involves replacing high-precision floating-point parameters with lower-precision integer parameters, thereby reducing the model's storage capacity and inference latency. Quantization methods can be categorized into Post-Training Quantization (PTQ) \cite{liu2021post} and Quantization-Aware Training (QAT)\cite{tailor2020degree} approaches.
At present, 4-bit quantization~\cite{Frantar_Ashkboos_Hoefler_Alistarh_2022} and mixed-precision quantization~\cite{Dettmers_Svirschevski_Egiazarian_Kuznedelev_Frantar_Ashkboos_Borzunov_Hoefler_Alistarh_2023} techniques are able to reduce memory cost of each weight element from 16-bit as a float point number to equivalently 3-4 bits.


In addition, Model sparsity pruning method mainly involves removing network elements by individual weights (unstructured pruning) or by entire rows and columns of weight matrices (structured pruning). Pruning can also be applied to various parts of the model, including entire layers\cite{sajjad2020poor}, heads, intermediate dimensions\cite{wang2019structured}, and blocks of weight matrices\cite{lagunas2021block}. The study of model pruning emerged since the 1980s' \cite{LeCun_Denker_Solla_1989,Hassibi_Stork_1992,Frantar_Alistarh_2022}, and it has demonstrated favorable outcomes across Computer Vision and NLP tasks. However, these methods require extensive retraining of pruned models to restore the performance level of the original models\cite{Ma_Fang_Wang}. While this is feasible for smaller-scale models, it becomes prohibitive for LLMs due to the associated computational costs. While there are some one-shot pruning methods that don't require retraining\cite{Hubara_Chmiel_Island_Banner_Naor_Soudry_2021, Frantar_Alistarh_2022}, their effectiveness are limited.
Therefore, extending the existing pruning methods to models with 10-100+ billion parameters is still an open and challenging problem. 

Recently, SparseGPT \cite{Frantar_Alistarh_2023} further developed the Optimal Brain Surgeon~(OBS)\cite{Hassibi_Stork_1992} approach. It achieved at least 50\% pruning ratio on LLMs with over a billion parameters in one shot, minimizing performance loss without requiring any retraining.
Based on previous studies in this area, we believe there is still considerable room for further improvements. Firstly, saliency criterion for weight element selection, such as OBS and OBD \cite{Hassibi_Stork_1992}, each captured one aspect of the weight element's saliency. We show that fusing the two criteria of OBS and OBD yields a better option for selecting the weights to be pruned, as observed in our experiments.


Moreover, SparseGPT assumes a uniform sparsity across all layers.
Inspired by the analysis of weight sensitivity in mixed-precision quantization methods \cite{Dong_Yao_Gholami_Mahoney_Keutzer_2019, Dong_Yao_Cai_Arfeen_Gholami_Mahoney_Keutzer_2019}, we investigated the sensitivity properties of different LLM layers. The magnitude of pruning ratios (sparsity) for each layer are then determined based on their respective sensitivities. Specifically, we utilize second-order information (the Hessian matrix) to assess the sensitivity of each layer.


Furthermore, we extend pruning to individual weight-level, an even finer granularity, for more precise trade-off. This means that each weight matrix is assigned a distinct sparsity level according to its sensitivity. In experiments with multiple cutting edge LLMs, our approach beats SparseGPT in terms of both perplexity and zero-shot downstream task performances.

In summary, our contributions are as follows: (1) We introduce a more comprehensive criterion for selecting the pruned weight elements. (2) We propose a novel sensitivity-aware mixed sparsity pruning strategy based on Hessian information. (3) To the best of our knowledge, this work achieves a new state-of-the-art performance in the field of LLM pruning.



\vspace{-1em}
\section{Methodology}
\label{sec:methodology}
\vspace{-0.5em}
This section describes the Improved Saliency Criterion (ISC) and the sensitivity-aware mixed sparsity pruning method.
ISC serves the purpose of selecting the optimal weight elements to be pruned, and the sensitivity level facilitates the allocation of the pruning ratios for different layers.

\subsection{Mask Selection and Weight Reconstruction Based on OBS}
\label{sec:methodology_mask}
In this sub-section, we review the OBS algorithm~\cite{Hassibi_Stork_1992} adopted in SparseGPT~\cite{Frantar_Alistarh_2023}, which serves as a part of our saliency criterion.
The OBS algorithm firstly prunes some weight elements, then updates the rest for error compensation.
Denoting the original weights as $W$, the objective is to ensure that the optimally pruned weights $W^*$ satisfy equation \eqref{eq:eq1}, where $X$ is the input data used for calibration purpose. 
\begin{equation}
\label{eq:eq1}
    W^* = \arg \min_{\hat{W}}||WX-\hat{W}X||_2^2,
\end{equation}

In the first step, to-be-pruned weight elements are determined through the saliency criterion, denoted \textit{e.g.}, $\varepsilon_m$ for the $m$-th row. In OBS, $\varepsilon_m$ is given by \eqref{eq:eq4}, where $w_m$ represents the original weight elements of the $m$-th row. $H^{-1}$ denotes the inverse matrix of $H = XX^T$ (the Hessian matrix of the objective function). One can select the weight elements associated with the smallest $p\%$ (sparsity level) of $\varepsilon_m$, and set them to 0. 

\begin{equation}
\label{eq:eq4}
   \varepsilon_m = \frac{w^2_m}{[H^{-1}]_{mm}}
\end{equation}

Secondly, one need to update the remaining weights to compensate for the error. Let $\delta W = W^* - W$ be the optimal adjustment of the remaining weight elements.
The objective then becomes minimizing the expectation of $||WX-(W+\delta W)X||_2^2$. Following Taylor expansion, the update formula is given by \eqref{eq:eq3} for the $m$-th row of $W$.

\begin{equation}
\label{eq:eq3}
    \delta W_m = - \frac{w_m}{[H^{-1}]_{mm}}·H^{-1}_{:,m}
\end{equation}

\subsection{Improved Saliency Criterion}
\label{sec:ISC}
An alternative saliency criterion $s$ had been proposed in the Optimal Brain Damage (OBD)~\cite{LeCun_Denker_Solla_1989} work, where $s = w_m^2H_{mm}$. The first step involves computing the second-order derivative of the loss with respect to the weights $W$, which is the Hessian matrix $H=XX^T$, derived from ~\eqref{eq:eq1}. The weight elements are then sorted according to the saliency levels, and elements with lower saliency are pruned. This approach aimed to achieve the minimal error in the pruning process.
However, it only focused on minimizing the direct error and did not take error compensation into account. 

Building upon the OBS and OBD approaches, this work considers combining the saliency criteria $\varepsilon$ and $s$, in order to balance the different aspects of previously used methods. We also apply error compensation following the approach in \ref{sec:methodology_mask}.
We define a more comprehensive saliency criterion -- ISC (Improved Saliency Criterion) in \eqref{eq:eq6}. This criterion allows for a more precise determination of pruning targets.

\begin{equation}
\label{eq:eq6}
    ISC =  w_m^2(H_{mm} + \frac{1}{[H^{-1}]_{mm}})
\end{equation}


\subsection{Mixed Sparsity Pruning Based on Hessian Sensitivity Awareness}
\label{sec:Mixed Sparsity Pruning Based on Hessian Sensitivity Aware}


In Section \ref{sec:methodology_mask} we assumed a constant sparsity ratio $p$. In this section, we propose a sensitivity-aware pruning strategy.

\vspace{-1em}
\subsubsection{Calculation of Sensitivity Based on Hessian Average Trace}
Sensitivity level is defined as the degree to which perturbations in model weights would affect the output error. Inspired by mixed-precision quantization~\cite{Dong_Yao_Cai_Arfeen_Gholami_Mahoney_Keutzer_2019}, we extend the sensitivity of model weights in mixed-precision quantization to the field of LLM pruning.
The sensitivity level is derived from the trace of the Hessian matrix $H$. Direct computation of $H$ for LLM is challenging because explicit calculation the Hessian matrix under cross-entropy loss would be time-and-space consuming. Nevertheless, we can address this issue using stochastic linear algebra\cite{Mahoney_2012} based methods for fast trace estimation. As shown in ~\eqref{eq:eq7}, $I$ denotes identity matrix, and $z$ is a random vector with \textit{i.i.d.} components, sampled from a standard Gaussian distribution. $N$ denotes the number of sampled data points. The second line in \eqref{eq:eq7} uses the Hutchinson algorithm~\cite{Avron_Toledo_2011} for efficient approximation, which we follow in the implementation.

\begin{equation}
\label{eq:eq7}
\begin{split}
    Tr(H) &=  Tr(HI) = Tr(HE[zz^T]) = E[Tr(Hzz^T)]  \\
    &=  E[z^THz] \approx \frac{1}{N}\sum_{i=1}^Nz_i^THz_i 
\end{split}
\end{equation}


For each weight matrix, we select the average trace of the Hessian as its sensitivity $sen$, as shown in \eqref{eq:eq9}, where $Tr(H)_{i,j}$ represents the trace of the Hessian matrix for the $j$-th weight matrix in the $i$-th layer, and $n$ denotes the number of diagonal elements in the Hessian matrix. $sen_{i,j}$ represents the sensitivity of the corresponding weight matrix.

\begin{equation}
\label{eq:eq9}
    sen_{i,j} = \frac{1}{n}Tr(H)_{i,j}
\end{equation}

\subsubsection{Rational Mixed Sparsity Assignment Based on Sensitivity}
\begin{figure}[t]
    \centering
    \vspace{-2em}
    \includegraphics[width=\linewidth, height=20em]{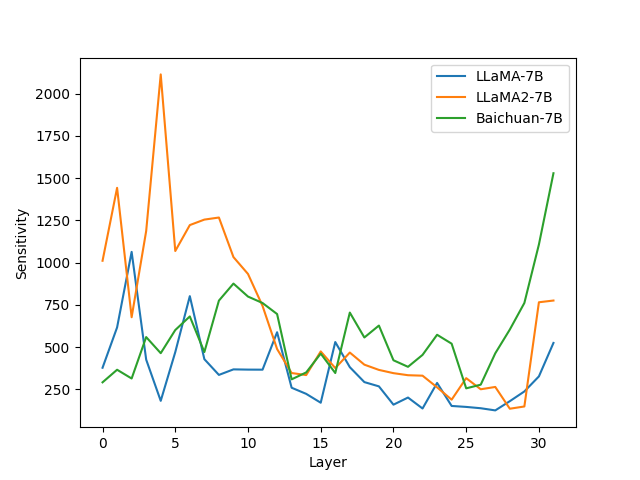}
    \caption{Sensitivity level of different layers for three LLM models: LLaMA-7B, LLaMA2-7B, and Baichuan-7B.}
    \label{fig:layer-level}
\end{figure}
\begin{figure}[t]
    \centering
    \includegraphics[width=0.95\linewidth,height=20em]{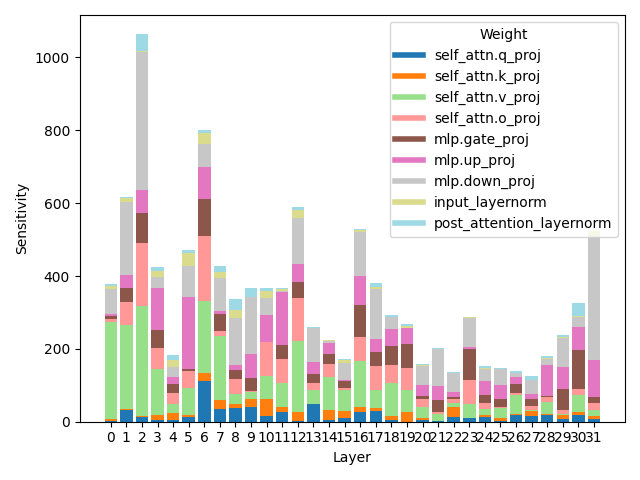}
    \caption{Sensitivity level of different weight components in LLaMA-7B.}
    \label{fig:weight-level}
\end{figure}
After obtaining the sensitivity level of weight matrices, we can allocate sparsity ratio individually to each one. This is referred to as \textbf{weight-level} mixed sparsity allocation. Pruning targets are selected adaptively within each weight matrix. We can also sum the sensitivities of all weight matrices in the same layer to obtain the sensitivity level of that layer. This is referred to as \textbf{layer-level} mixed sparsity allocation.

Layer-level sensitivity analysis can be seen in Figure \ref{fig:layer-level}. The three different models exhibit some commonalities in that the bottom and top layers are more likely to show higher sensitivities, while the intermediate layers have lower sensitivities. This may be attributed to the fact that the bottom layers determine the model's initial encoding of the input embeddings, and their variations can have a significant impact on the intermediate layers, thereby affecting the overall model performance. The top layers are also sensitive since they have a direct influence on the model's output.
Weight-level sensitivity analysis can be seen in Figure \ref{fig:weight-level}. The value matrices (plot in green color) in self-attention modules and the down projection weights (plot in gray color) in MLP modules are relatively more sensitive.

We use Algorithm~\ref{alg:SABS} to determine the sparsity of 
each unit (layer or weight) based on their sensitivities. We rank the sensitivity of the weight matrices, then assign a corresponding sparsity to each one, ensuring that the overall sparsity of this mixed sparsity approach equals a pre-defined overall sparsity level $p$.
For example, the typical value for sparsity interval's half-width $\alpha$ is 0.1.
When the overall sparsity $p$ is 0.5, L and R would be 0.4 and 0.6, respectively. This means that the sparsity of each weight falls within the interval [0.4, 0.6], and the total sparsity of all weights adds up to 0.5. 

\begin{algorithm}
   \caption{Sparsity Assignment Based on Sensitivity}
     \label{alg:SABS}
     \textbf{Input:} $sen$, The sensitivity of each weight within each layer.\\
     \textbf{Input:} Overall sparsity $p$, sparsity interval's half-width $\alpha$.\\
     \textbf{Output:} $spa$, The sparsity of each weight within each layer. 
     \begin{algorithmic}[1]
     \Function{Sparsity-Assignment}{$sen$,$p$,$\alpha$} 
     \Let{total\_weight}{[]} 
            \State Append(all elements of $sen$) to total\_weight
     \Let{total\_weight}{sorted(total\_weight)}
     \For{$i \gets 0 \textrm{ to } \text{length($sen$)-1}$}
        \For{$j \gets 0 \textrm{ to } \text{length(sen[i])-1}$} 
            \Let{id}{rank(total\_weight, $sen$[i,j])} \# base 0
            \Let{L}{1 - $p$ - $\alpha$}
            \Let{R}{2 * (1 - $p$) - L}
            \Let{$spa$[i,j]}{L + id * (R - L) / (len(total\_weight) - 1)} 
        \EndFor
     \EndFor
     \State \Return $spa$
     \EndFunction  
    \end{algorithmic}
\end{algorithm}


\section{Experiments}
\label{sec:experiments}

\vspace{-1em}
\subsection{Experiment Setup}
\label{sec:exp_setup}
\subsubsection{Model, Dataset and Evaluation}
We employ several cutting edge LLM models in the experiments, including LLaMA~\cite{touvron2023llama}, LLaMA2~\cite{touvron2023llama2}, and Baichuan \cite{baichuan_website_2023}. We prioritize the study on 7B/13B models as these are more widely adopted in research and applications. We evaluate the pruned models on standard benchmarks  such as raw-WikiText2~\cite{WikiText2}, PTB~\cite{PTB}, and C4~\cite{C4} (a subset of the validation sets). As usual, we employ perplexity as the evaluation metric.


To examine the pruned models' performances on downstream tasks, we conduct zero-shot accuracy tests on the OpenLLM LeaderBoard\footnote{\url{https://huggingface.co/spaces/HuggingFaceH4/open_llm_leaderboard}}
by utilizing the popular open-source auto-regressive language model evaluation framework EleutherAI-evalharness
\footnote{\url{https://github.com/EleutherAI/lm-evaluation-harness}}
.
We employ a variety of tasks including ARC (Challenge)~\cite{ARC}, HellaSwag~\cite{zellers2019hellaswag}, TruthfulQA (Multiple-Choice2)~\cite{lin2021truthfulqa}, and PIQA~\cite{Tata_Patel_2003} to compare the pruned models with their corresponding original versions. This allows us to assess the performance of the model across a variety of tasks and gain insights into its generalization capabilities. The experimental results of our approach are primarily compared with the relative values of dense and baseline models. 

\vspace{-1em}
\subsubsection{Setup}
We implement our method using PyTorch and HuggingFace's Transformers library.
The most computation-heavy step in our experiment was around calculating the trace of the Hessian matrix, which required 8 NVIDIA A100 GPUs for 13B-parameter models.
All pruning experiments are one-shot and do not require any fine-tuning or retraining.
For calibration data, we utilized 128 2048-token segments. These segments are randomly selected from the first partition of the C4 dataset. We ensure that no task-specific data is seen during the pruning process to guarantee zero-shot conditions.
The baselines we compare against mainly consist of the standard magnitude pruning ~\cite{Zhu_Gupta_2017} and SparseGPT~\cite{Frantar_Alistarh_2023} methods.

\subsection{Experiment Results and Analysis}
\label{sec:exp_results}
\subsubsection{Evaluation on Improved Saliency Criterion}
In this experiment, we verify the effectiveness of the proposed saliency criterion. In Table~\ref{tab:ISC}, we compare the perplexity for the two LLM models under different sparsity strategies and saliency criteria. The experiment results have validated the effectiveness of ISC as well as the mixed sparsity strategy, as the perplexity of the pruned models were lower in comparison to Uniform sparsity and OBS saliency on all of the examined data sets.

\begin{table}[h]
    \centering
    \caption{Performance perplexity comparison under the condition of an overall sparsity level of 50\%. ``Uniform'' denotes a uniform sparsity level; ``Mixed'' denotes the proposed sensitivity-aware pruning strategy.}
    \begin{adjustbox}{width=0.5\textwidth}
    \begin{tabular}{l|l|c|c|c|c}
        \toprule
        Model & Sparsity & Saliency & WikiText2$\downarrow$ & PTB$\downarrow$ & C4$\downarrow$ \\
        \midrule
        
        \multirow{4}{*}{LLaMA7B} & \multirow{2}{*}{Uniform} & OBS & 7.79 & 65.77 & 10.17 \\
        
        & & ISC & 7.11 & 61.42 & 9.27 \\
        
        & \multirow{2}{*}{Mixed} & OBS & 6.77 & 52.10 & 8.88 \\
        
        & & ISC & \textbf{6.76} & \textbf{50.12} & \textbf{8.82} \\

        \midrule
        \multirow{4}{*}{Baichuan7B} & \multirow{2}{*}{Uniform} & OBS & 16.74 & 43.82 & 27.43 \\
        
        & & ISC & 14.77 & 38.54 & 24.54 \\
       
        & \multirow{2}{*}{Mixed} & OBS & 14.43 & 38.50 & 24.13 \\
        
        & & ISC & \textbf{12.29} & \textbf{33.04} & \textbf{20.63} \\
        \bottomrule
    \end{tabular}
    \end{adjustbox}
    \label{tab:ISC}
\end{table}
\vspace{-1em}

\subsubsection{Evaluation on Mixed Sparsity Pruning}
We adopt a mixed sparsity approach to prune different base models, as shown in Table~\ref{tab:MSP}. It can be seen that our method outperforms SparseGPT in three mainstream 13B LLM models. The experiment results demonstrate that our method can further reduce the performance degradation caused by pruning, bringing it closer to the dense model. Since the overall sparsity is set at 50\%, our method and SparseGPT achieve an equal model compression ratio and inference acceleration effect.

Moreover, we conduct tests on zero-shot downstream NLP tasks. As shown in Table~\ref{tab:zero_shot}, with an overall sparsity level of 50\%, the proposed approach outperforms SparseGPT in terms of accuracy. These results validate that our method is effective in reducing the errors introduced by pruning.

\begin{table}[h]
    \centering
    \caption{
Under the condition of an overall sparsity rate of 50\%, the three base models have different perplexity results under various methods.}
    \begin{tabular}{l|l|c|c|c}
        \toprule
        
        Model & Method & WikiText2$\downarrow$ & PTB$\downarrow$ & C4$\downarrow$ \\
        
        \midrule
        \multirow{4}{*}{LLaMA-13B} & Dense & 5.03 & 32.26 & 6.80 \\
        
        & Magnitude & 20.25 & 149.43 & 25.11 \\
        
        & SparseGPT & 6.12 & 43.92 & 8.13 \\
        
        & Ours & \textbf{5.68} & \textbf{38.69} & \textbf{7.56} \\
        \midrule
        \multirow{4}{*}{LLaMA2-13B} & Dense & 4.88 & 40.99 & 6.73 \\
        
        & Magnitude & 6.77 & 117.69 & 9.38 \\
        
        & SparseGPT & 5.99 & 59.03 & 8.23 \\
        
        & Ours & \textbf{5.51} & \textbf{53.03} & \textbf{7.57} \\
        \midrule
        \multirow{4}{*}{Baichuan-13B} & Dense & 5.61 & 16.49 & 8.23 \\
        
        & Magnitude & 15.64 & 48.74 & 22.23 \\
        
        & SparseGPT & 7.52 & 21.46 & 10.95 \\
        
        & Ours & \textbf{6.63} & \textbf{19.25} & \textbf{9.71} \\
        \bottomrule
    \end{tabular}
    \label{tab:MSP}
\end{table}

\begin{table}[h]
    \centering
    \caption{Comparison of the accuracy results of the different methods for zero-shot with an overall sparsity of 50\%, where HS stands for HellaSwag and TQ stands for TruthfulQA.}
    \begin{adjustbox}{width=0.5\textwidth}
    \begin{tabular}{l|l|c|c|c|c|c}
        \toprule
        Model & Method & ARC$\uparrow$ & HS$\uparrow$ & TQ$\uparrow$ & PIQA$\uparrow$ & Avg$\uparrow$ \\
        \midrule
        \multirow{3}{*}{LLaMA-7B} & Dense & 51.0 & 77.8 & 34.3 & 79.2 & 60.6 \\
        
        & SparseGPT & 39.5 & 69.2 & 36.2 & 76.6 & 55.4  \\
        
        & Ours & \textbf{40.2} & \textbf{71.2} & \textbf{38.1} & \textbf{77.5} & \textbf{56.8}\\
        
        \midrule
        \multirow{3}{*}{LLaMA-13B} & Dense & 47.6 & 79.1 & 39.9 & 80.1 & 61.7 \\
        
        & SparseGPT & 41.8 & 73.9 & 37.5 & 79.0 & 58.1 \\
        
        & Ours & \textbf{43.8} & \textbf{76.5} & \textbf{40.4} & \textbf{79.8} & \textbf{60.1} \\
        \midrule
        \multirow{3}{*}{LLaMA2-7B} & Dense & 46.3 & 75.9 & 38.9 & 79.1 & 60.1  \\
        
        & SparseGPT & 41.3 & 71.1 & 38.3 & 76.9 & 56.9 \\
        
        & Ours & \textbf{44.1} & \textbf{72.8} & \textbf{39.4} & \textbf{77.5} & \textbf{58.5}\\
        
        \midrule
        \multirow{3}{*}{LLaMA2-13B} & Dense & 48.9 & 79.4 & 36.9 & 80.5 & 61.4  \\
        
        & SparseGPT & 44.9 & 74.3 & 37.9 & 77.9 & 58.8  \\
        
        & Ours & \textbf{47.5} & \textbf{76.9} & \textbf{39.4} & \textbf{79.5} & \textbf{60.8}\\
        \midrule
        
        \bottomrule
    \end{tabular}
    \end{adjustbox}
    \label{tab:zero_shot}
\end{table}

\vspace{-2em}
\subsubsection{Evaluation on Very High Sparsity Models}
\begin{figure}
    \vspace{-1em}
    \centering
    \includegraphics[width=\linewidth, height=20em]{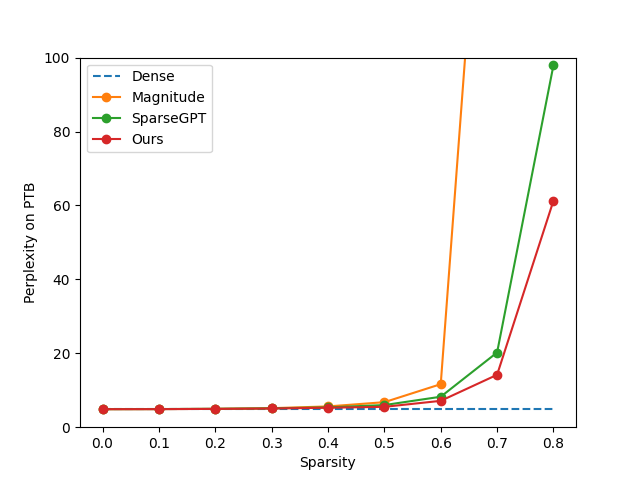}
    \caption{Perplexity of different methods with varying sparsity levels evaluated on PTB.}
    \vspace{-0.5em}
    \label{fig:ppl}
\end{figure}

Figure~\ref{fig:ppl} shows that performance degradation expands when the sparsity level increases, which is a common effect for all pruning methods examined (results for LLaMA2-13B model plotted). Nevertheless, the proposed method helps mitigating the degradation even more as the sparsity level becomes higher.

\vspace{-0.5em}
\subsubsection{Joint Mixed Sparsity Pruning and Quantization}

In this experiment, we carry out joint quantization and pruning of the model, to see if their effects can be combined. We follow the approach of SparseGPT, where pruning is applied prior to quantization. As shown in Table~\ref{tab:Joint Pruning and Quantization}, 3-bit quantization using GPTQ method yields much heavier loss in comparison to 4-bit quantization, with only minor increase of compression ratio~\cite{Frantar_Ashkboos_Hoefler_Alistarh_2022}. However, when we combine SparseGPT with 50\% sparsity and 4-bit quantization, higher compression ratio and lower performance loss are achieved together. Moreover, as indicated in the last row, the proposed method further improves the performance compared to SparseGPT at the same compression ratio, confirming the effectiveness of the mixed sparsity approach. It offers a new SOTA performance via combining quantization and pruning.

\begin{table}[h]
    \centering
    \caption{The perplexity performance of LLaMA-7B after compression under different methods. \texttt{Ratio} represents the theoretical model compression ratio.}
    \begin{adjustbox}{width=0.5\textwidth}
    \begin{tabular}{l|c|c|c|c}
        \toprule
        Method & WikiText2$\downarrow$ & PTB$\downarrow$ & C4$\downarrow$ & Ratio$\uparrow$\\
        \midrule
        Dense & 5.63 & 35.79 & 7.34 & 1.0 \\
        
        4bit GPTQ~\cite{Frantar_Ashkboos_Hoefler_Alistarh_2022} & 7.07 & 34.34 & 8.19 & 4.0$\times$ \\
        
        3bit GPTQ & 35.25 & 226.29 & 17.92 & 5.3$\times$\\
        SparseGPT 50\% + 4bit GPTQ & 12.80 & 88.60 & 11.37 & \textbf{8.0$\times$} \\
        Ours 50\% + 4bit GPTQ & \textbf{10.23} & \textbf{65.49} & \textbf{9.91} & \textbf{8.0$\times$} \\
        \bottomrule
    \end{tabular}
    \end{adjustbox}
    \label{tab:Joint Pruning and Quantization}
\end{table}


\vspace{-1em}
\section{Conclusions}
\label{sec:conclusions}
In this paper, we have proposed a mixed sparsity pruning method based on Hessian sensitivity awareness, which assigns sparsity levels to each weight by discerning their sensitivity through the average trace of the Hessian matrix. Under the same overall sparsity constraint as SparseGPT, our mixed sparsity allocation scheme further reduces the error introduced by pruning. Moreover, our method exhibits even more significant advantages in scenarios with extremely high overall sparsity. Additionally, our approach is compatible with quantization, achieving higher compression ratios while minimizing model performance degradation. Future work will explore the combination of our approach with mixed-precision quantization to achieve further improved results.

\section{Acknowledgements}
\label{sec:acknowledgement}
\vspace{-0.5em}
This work was supported in part by China NSFC projects under Grants 62122050 and 62071288, in part by Shanghai Municipal Science and Technology Commission Project under Grant 2021SHZDZX0102.

\bibliographystyle{IEEEtran}
\bibliography{refs}

\end{document}